\documentclass[12pt,english]{scrartcl}
\usepackage[T1]{fontenc}
\usepackage{dfadobe}  

\usepackage[backend=biber,bibstyle=EG,citestyle=alphabetic,backref=true]{biblatex} 
\ifpdf \usepackage[pdftex]{graphicx} \pdfcompresslevel=9
\else \usepackage[dvips]{graphicx} \fi

\usepackage{array}
\usepackage{soul}
\usepackage{authblk}

\usepackage{egweblnk} 

\usepackage[colorinlistoftodos,prependcaption]{todonotes}
\definecolor{turquoise}{cmyk}{0.65,0,0.1,0.1}
\definecolor{purple}{rgb}{0.65,0,0.65}
\definecolor{dark_green}{rgb}{0, 0.4, 0}
\definecolor{dark_blue}{rgb}{0, 0, 0.4}
\definecolor{orange}{rgb}{0.6, 0.3, 0.0}
\definecolor{red}{rgb}{0.8, 0.2, 0.2}
\definecolor{brown}{rgb}{0.5, 0.16, 0.16}

\newif\ifshowauthorcomments

\showauthorcommentstrue

\newif\ifshowedits

\showeditsfalse

\newcommand{\hle}[1]{\ifshowedits{\hl{#1}}\else{#1}\fi}

\newcommand{\defTodo}[2]{%
  \expandafter\newcommand\csname #1\endcsname[1]{%
    \todo[linecolor=#2,backgroundcolor=#2!25,bordercolor=#2,inline]{\textbf{#1}: ##1}}}

\newcommand{\defTODO}[2]{%
  \expandafter\newcommand\csname #1\endcsname[1]{%
    \todo[linecolor=#2,backgroundcolor=#2!25,bordercolor=#2,inline,caption={\textbf{(#1 LONG TODO)}}]{##1}}}

\newcommand{\deftodo}[3]{%
  \defTodo{#1}{#3} \defTODO{#2}{#3}}

\deftodo{Chandra}{CHANDRA}{Magenta}
\deftodo{Adam}{ADAM}{RoyalBlue}
\deftodo{Molly}{MOLLY}{RoyalBlue}
\deftodo{Dan}{DAN}{RoyalBlue}
\deftodo{Eva}{EVA}{RoyalBlue}
\deftodo{Zach}{ZACH}{Plum}

\newcommand{\pointcloud}{\mathcal{S}}

\newcommand{\parts}{\mathcal{P}}
\newcommand{\parti}{P}

\newcommand{\pose}{T}
\newcommand{\depth}{d}
\newcommand{\shape}{\Theta}

\newcommand{\shapeinitial}{\Theta^0}

\newcommand{\adjacency}{\mathrm{Adj}}
\newcommand{\diameter}{D}

\newcommand{\mask}{\mathbf{M}}
\newcommand{\coordinates}{\mathcal{V}}
\newcommand{\edges}{\mathcal{E}}
\newcommand{\rectifiedimage}{\tilde{\mathbf{I}}}

\newcommand{\connectionthreshold}{\tau_c}
\newcommand{\penetrationthreshold}{\tau_c}
\newcommand{\occlusionmask}{\mathbf{V}}
\newcommand{\plane}{\pi}
\newcommand{\minangle}{\alpha_{\mathrm{min}}}

\begin{document}

\title{Fabrication-Aware Reverse Engineering for Carpentry}

\author{\parbox{\textwidth}{\centering James Noeckel$^{1}$, 
        Haisen Zhao$^{1,2}$, \\ 
        Brian Curless$^{1}$, and 
        Adriana Schulz$^{1}$
        }
}

\affil{$^1$University of Washington\\
         $^2$Shandong University}
         
\date{Accepted for publication in the Eurographics Symposium on Geometry Processing 2021}

\maketitle

\begin{abstract}
We propose a novel method to generate fabrication blueprints from images of carpentered items. While 3D reconstruction from images is a well-studied problem, typical approaches produce representations that are ill-suited for computer-aided design and fabrication applications. Our key insight is that fabrication processes define and constrain the design space for carpentered objects, and can be leveraged to develop novel reconstruction methods. Our method makes use of domain-specific constraints to recover not just valid geometry, but a semantically valid assembly of parts, using a combination of image-based and geometric optimization techniques. We demonstrate our method on a variety of wooden objects and furniture, and show that we can automatically obtain designs that are both easy to edit and accurate recreations of the ground truth. We further illustrate how our method can be used to fabricate a physical replica of the captured object as well as a customized version, which can be produced by directly editing the reconstructed model in CAD software. 
\end{abstract}

\begin{figure*}
  \centering
  \includegraphics[width=\textwidth]{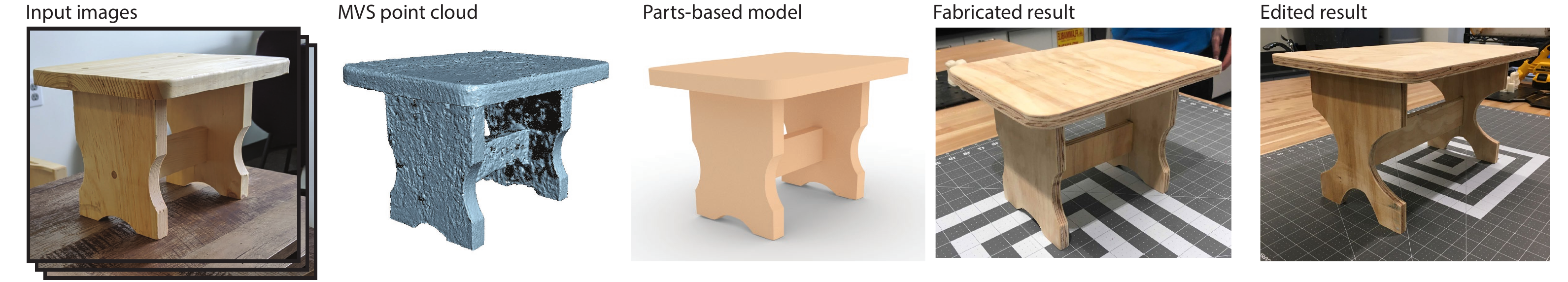}
  \caption{Given a set of images of a carpentered object and the resulting multi-view stereo (MVS) reconstruction, we 
  produce a concise, part-based assembly that can be fabricated, producing a physical replica nearly identical to the original.
  Our model is easily edited in CAD software, allowing for the fabrication of modified versions of captured designs (right).}
  \label{fig:teaser}
\end{figure*}

\section{Introduction}

\label{sec:intro}



Carpentered, wooden furniture and objects abound in our everyday lives, from stools to bookshelves to trays for carrying food.  Imagine though you find a piece you like at a friend's house, but you \hle{are unable to} find it for purchase, or perhaps you even want a slightly different version of it, wider or with changes to the curves in its shape.  Suppose then you could simply take some photos of the piece, run some software, and generate a CAD model that represents how the object was made---one that is easy to edit, if desired---and then build just by cutting parts out of sheets of wood and assembling them, which you could even do yourself.  In this paper, we propose to tackle just this problem, taking as input a set of images of a carpentered object and generating a CAD model that is ready for building a replica of the object, or an edited version of it.


Typical methods for object capture from images perform structure-from-motion to recover camera poses and sparse scene points, followed by multi-view stereo to densify the reconstruction. The result is usually a point cloud or, with some processing, a dense triangle mesh, exhibiting noise and incomplete coverage.  These representations are far from being complete, concise, editable CAD models and further tell us nothing about how to cut and assemble parts to build the model. 

In this work, we propose a novel direction for recovering representations of fabricated objects: describing the \textit{fabrication process itself}. By framing this as a reverse engineering problem in a specific fabrication domain, we introduce constraints that significantly reduce the search space of viable 3D models.
In particular, we operate on carpentered objects consisting of parts that are cut from sheets of wood and then connected together. The space of fabrication instructions in this domain is still highly expressive, covering a variety of everyday objects, as we will show, while also adhering to the real-world constraints governing the construction process, so that the output is by definition ready to build.



This reverse engineering problem introduces its own set of challenges: arriving at a fabricable solution requires identifying the parts, and optimizing for their precise shapes and the part-to-part connections constraining those shapes.  This mixture of discrete and continuous degrees of freedom makes for a challenging optimization problem; to make this more tractable, we propose a multi-stage algorithm in which we first select the initial geometry and positions of parts in the assembled object, progressively detect assembly constraints (i.e. connections between parts), and then refine the geometry subject to these new constraints.  The input images, captured by simply walking around an object and taking photos with a smartphone, guide this process at multiple stages.  First, we use the images to recover a multi-view stereo point cloud that, though incomplete, drives the initial CAD part recovery.  Second, the images provide evidence of seams -- discontinuities in appearance -- that indicate how different pieces of wood fit together when the connections are otherwise ambiguous based on geometry alone.  Finally, by rectifying the images to each part plane, we can co-segment the part faces to obtain more accurate contours, i.e., cut paths for fabrication. Each of these co-segmented contours, extracted at the pixel level, is not concise and may not respect assembly constraints; we additionally propose an algorithm to find a simple parametric boundary that accurately represents the cut path of each part while respecting contact constraints between parts.

Our key contributions are:
\begin{itemize}
    \item A fabrication-aware pipeline for selecting a plausible part structure representing the input object
    \item An algorithm for recovering contacts between connected parts using geometric and image evidence
    \item A method for extracting cut paths representing part shapes using multiple image views
    \item A method for incorporating assembly constraints into fitting of regularized, parametric contours to imperfect data
\end{itemize}

We note that, since our approach is based on features observed in images, we assume that the carpentered objects are textured, which is typical for wood that is unfinished or varnished, but not painted a uniform color.  Further, though we don’t require complete reconstruction of the surface, we do require that the wood sheet surface of each part be at least partially visible.  Finally, we restrict the class of objects reconstructed to those that can be assembled with parts cut from sheets of wood.
We demonstrate results on a variety of objects of different size and complexity, and show the efficacy of our method by fabricating two of our results, along with edited versions.

\section{Related Work}
\label{sec:related}

Reverse Engineering (RE) aims to recover CAD models from measured data \cite{buonamici2018reverse}, concisely representing the various parts and geometric features using parametric primitives and surfaces. RE methods can be labeled according to their target CAD representation, which contain varying levels of structure. We can classify these as \textbf{low-level} (volumetric models \cite{chivate1993solid}, meshes \cite{xu2011photo}), \textbf{medium-level} (primitives \cite{li2019supervised}, B-Rep surfaces \cite{benkHo2001algorithms}, surface patch-based representations \cite{eck1996automatic}, procedural shape structures \cite{du2018inversecsg}) and \textbf{high-level} (parametric CAD models \cite{smirnov2019deep}, and multi-component 3D models with geometric constraints \cite{xu2016interactive,Lau2011FabPartsConnectors}). The distinction we make between mid- and high-level representations is that high-level contains both full geometric information and some additional structure relating to the semantics of the object, such as shape parameters corresponding to degrees of freedom in the design and relationships between assembled 3D parts. While Mid-level representations such as primitives, B-Reps, surface patches, and CSG trees may capture the geometry more concisely than low-level, they do not inherently encode these global semantics. 
Since our target is a CAD model \hle{complete with parts and connections} that would allow us to physically reproduce the object, we focus on high-level representations. \hle{For a survey of works related to optimizing rigid assemblies of parts, see} \textcite{wang2021stateoftheart}.

\paragraph*{Retrieval-based Reconstruction} One approach to reverse engineering is to retrieve plausible objects from a shape library and align them with the input. \textcite{Avetisyan2019CVPR} and \textcite{lim2013parsing} detect and align representative objects from a known database with an input 3D scan or image, respectively. \textcite{schulz2017retrieval} and \textcite{uy2020deformation} proposed strategies for deformation-aware retrieval of single CAD shapes.  \textcite{xu2011photo} present an interactive method for using predefined 3D models as templates to deform to match an input photograph, and \textcite{huang2015single} retrieves individual parts from a small 3D model database to reverse engineer the object from a single image. Because these methods rely on a database containing examples that sufficiently resemble the query object, they do not generalize to unseen classes of objects. While we recognize the need to limit the scope to ensure that the problem is tractable, we wish to do this by restricting the fabrication process instead of the types of allowed objects, which permits a variety of input beyond what can be memorized in a database, so retrieval-based methods are not suited to our task.

\paragraph*{Classical Reverse Engineering}
Obtaining CAD models from measured data is a well-studied problem. Many works concerned with reconstructing CAD models follow a similar strategy, with the common features being segmentation of the input point cloud/mesh, fitting analytical surfaces to the segmented regions, and stitching the disjoint surfaces into a complete CAD model \parencite{buonamici2018reverse, reverseeng2012}. The goal of these methods is accurate recovery of CAD surface primitives, which lack the high-level assembly information that we require. 
Nevertheless, the techniques of feature-based reverse-engineering are widely useful tools for inferring geometry from dense 3D data; we build our model in part based on detected plane and cylinder features.


\paragraph*{Interactive Reverse Engineering}
In solving the difficult inverse problem of reverse engineering, some works allow for user interaction to resolve ambiguities and provide hints for reconstruction. \textcite{chen20133} recovers interactive manipulable 3D shapes from a single photograph, guided by user-supplied sketches of generalized cylinders and other primitives. \textcite{xu2016interactive} recovers dynamic mechanical structures from a single image, where the user sketches part profiles and supplies hints for how detected parts should snap to their surroundings. \textcite{xu2011photo}, which deforms retrieved models to match a target image, relies on the user to aid in semantic segmentation of the input image, as well as for selecting the candidate shape to deform. \textcite{arikan2013osnap} \hle{reconstructs architectural models from point clouds by first automatically fitting a coarse 3D model, then using a sketch-based interface to allow users to add additional geometric details while optimizing the consistency and accuracy of the model.
} Our proposed reverse engineering method, once provided with a 3D reconstruction of the object, is fully automatic.

\paragraph*{Learning-based Reverse Engineering}
Some existing works infer high-level abstract 3D structures from images \parencite{niu2018im2struct}, 3d meshes \cite{tulsiani2017learning} and shape structures \parencite{jones2020Assembly}. In these works, the output structures are represented using coarse cuboids, which fall short of our goal of a full part-based CAD representation, while some require segmented part hierarchies as input \parencite{jones2020Assembly}. In \parencite{ganapathi2018parsing}, hand-crafted abstract structures, consisting of a tree of axis-aligned cuboids and connectors, are fitted to incomplete 3D scans to aid in classification and shape completion. Inferred abstract structures can be used to aid in retrieval of complete CAD models, but this again is limited by what is recorded in a database, along with a learned template for the corresponding object class. In a recent work, \textcite{smirnov2019deep} introduce a unified learning framework in which parametric 2D and 3D primitives can be inferred from raster or voxel representations. Using this framework, it is possible to learn primitives in 2D and 3D that encode certain semantics of reconstructed objects, for example particular CSG primitives might correspond to the armrests of chairs. However, as with many learning-based works, these semantics are specific to the class of objects on which the method is trained, and for each such class, a template encoding these semantics is a prerequisite for training. If we wished to reconstruct a unique type of object whose only familiar feature is the carpentry construction process, it would be difficult to acquire the data to produce meaningful shape and structure predictions using any of the aforementioned learning- or retrieval-based methods. Conversely, our method can recover the shape and structure of such an unknown object, as we only assume that objects obey the geometric constraints imposed by carpentry. 

\paragraph*{Grammar- and program-based Reverse Engineering}
A number of domain-specific reverse engineering methods exist which utilize known structures and semantics of a narrow class of objects. \cite{Lau2011FabPartsConnectors} infers a rich fabricable design consisting of parts and connectors, but it does not generalize beyond a few furniture classes for which specific hand-crafted grammars are defined. \cite{fan2016probabilistic} models the exterior of residential buildings by learning a probabilistic model from street-view images. This enables them to infer plausible building geometry using (potentially occluded) single views as input, but it is inherently limited in how accurately it can model the input due to its non-determinism and the lack of additional evidence to constrain the result. 
\cite{tian2019learning} Presents a domain-specific language and neural program executor for learning to synthesize CAD models, with the ability to represent symmetries and repeated structures through programmatic loops. Their neural program executor allows for differentiable rendering of varying-length programs, facilitating unsupervised fine-tuning on unannotated shapes, which helps the method to generalize beyond training categories. However, the design of the DSL incorporates furniture-specific semantic annotations, so it is still limited to classes of objects where such labels apply. 

In the carpentry domain, grammars have also been used outside of reverse engineering tasks, including interactive design systems~\cite{Umetani:GuidedExploration,koo2014creating,Song2017reconfigurableFurniture, garg2016computational,Fu:InterlockingFurniture} and optimization of fabrication plans~\cite{yang2015reforming,Koo:ZeroWasteFurniture,jigfab,Lau2011FabPartsConnectors, wu2019carpentry}. Our work uses similar fabrication-aware representations for reverse-engineering.

\cite{Lin2018RecoveringFM} solves a different but related problem, and employs similar ideas to ours for reverse engineering functional mechanical assemblies from raw scans, in that their general approach consists of detecting parts from input scans, determining interactions between parts, and globally optimizing the model geometry to satisfy the resulting constraints. However, the parts they consider are limited to a set of predefined templates suited to their domain.




\paragraph*{Decomposition-based Methods}
\hle{The goal of these methods is to convert a 3D surface representation into solid, fabricable parts that recreate the desired surface} \parencite{araujo2019surface2volume,filoscia2020optimizing,yao2017interactive}. \hle{A related method is to segment structured 3D models into repeated sub-components}  \parencite{demir2015coupled}. \hle{However, these methods do not directly apply to the problem of reverse engineering from images.}

\paragraph*{Model-based Reconstruction from Images}
Many works in reconstruction from images seek to make use of model-based assumptions, dating back to early ``blocks world'' work for recovering 3D edges in an image ~\parencite{roberts63blocks}. A large body of work in model-based reconstruction from images has focused on architecture, making use of vanishing points  ~\parencite{criminisi2000single} and repeated structures such as windows ~\parencite{dick2004modelling} that can be exploited for geometric information. This has given rise to interactive modeling systems for recovering building facades ~\parencite{sinha2008interactive, facades}, as well as fully automatic systems \parencite{werner2002new}. Such systems typically exploit architectural assumptions of abutting cuboids sitting on the ground and sloped roofs which hold for the structures they model, while we seek to accurately reconstruct individual, solid parts of objects with arbitrary curved shapes in potentially complex arrangements.

%

\begin{figure*}
    \centering
    \includegraphics[width=150mm]{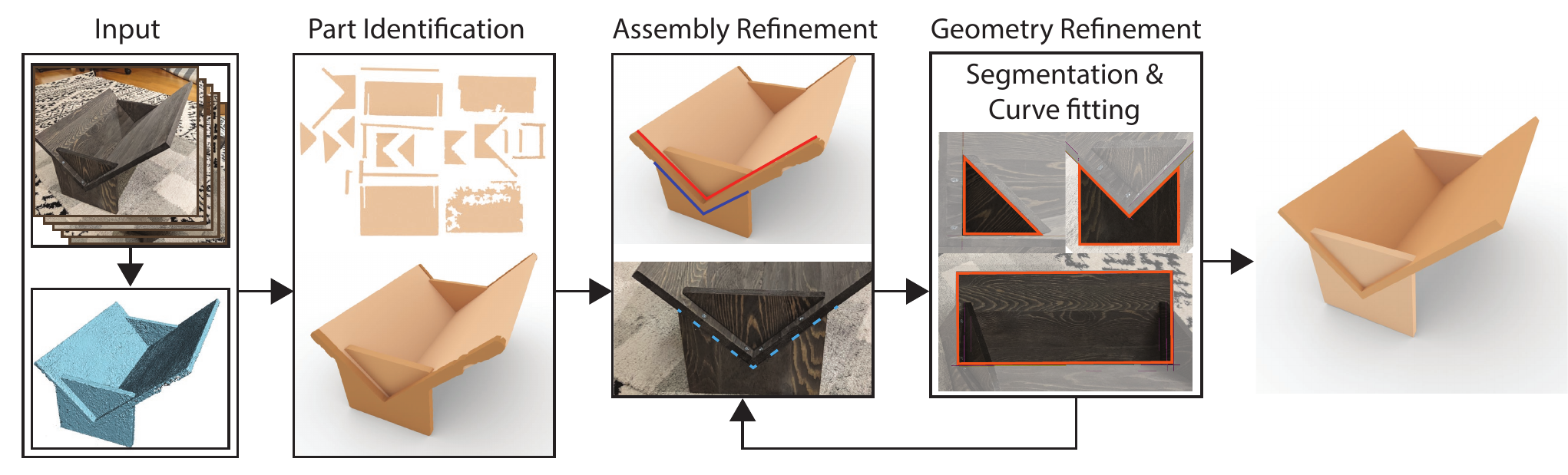}
    \caption{Overview of our method: Given input images, we recover an oriented point cloud with 3D reconstruction software. The images, camera poses, and point cloud are inputs to our system.  We then identify an overcomplete set of parts based on plane fits to the point cloud, along with the part's cut path and thickness. These are pruned to minimize volume overlap while still adhering to the point cloud. The resulting part set is then refined in the assembly stage that determines how parts connect, resolves remaining volume conflicts (based on visible evidence of seams), and re-aligns the parts. The geometry refinement stage then uses multi-view image segmentation followed by concise curve fitting, both constrained by part connections, to obtain more accurate and more CAD-like cut paths. The geometry refinement can lead to cut paths that imply new part contacts, and thus we alternate between assembly and geometry refinement until the final model is complete.
    }
    \label{fig:overviewfig}
\end{figure*}



\section{Overview}

The input to our algorithm is a set of images of a carpentered object taken from different viewpoints. The output is a fabricable model 
describing the set of parts along with how they should be connected. Parts are assumed to be cut from wood \textit{sheets}, so that their boundary contains two \textit{sheet planes} (from the front and back of the sheet). Parts are represented 
as a triplet of the wood sheet position $\pose$ (a rigid transformation), the sheet thickness $\depth$, and cut path $\shape$ represented in the 2D plane of the sheet. We allow individual straight cuts to be made at arbitrary angles against the wood plane to allow for slanted contacts between parts; $\shape$ also contains these bevel angles. 
Note that the cut path may include holes in the interior of the shape. 
The assembly information specifies a list of part pairs that are joined, and the surfaces along which these connections occur, which we represent using \textit{interfaces} that we assume to be planar, as planar contacts are common in carpentry assembly. We say the model is \textit{fabricable} if, in their assembled configuration, the parts do not overlap and meet snugly at the joints. We refer to the pairwise contact constraints implied by this requirement as \textit{assembly constraints}.

While assembly constraints assist in recovering partially unobserved parts, the need to infer both the set of parts and the assembly constraints from incomplete data presents a challenge: Inferring the shapes and positions of parts depends on how they are connected, and likewise finding probable connections depends on the part geometry. This interdependence implies that these properties should be considered jointly in order to arrive at a feasible solution. To address the complex search space of possible part assemblies, we adopt a multi-stage approach in which we first detect an approximate set of parts absent any connections, then iteratively refine the model by alternately optimizing for the connection contacts and the part geometry subject to the new contact constraints. Finally, we approximate each cut path with a concise, piecewise-smooth curve that balances simplicity and accuracy.
Our approach is illustrated in Figure \ref{fig:overviewfig}.
 
\paragraph*{Preparing the Input.}
Given our images, we use 3D reconstruction software ~\cite{capturingreality} to obtain camera poses and a semi-dense, oriented, point cloud $\pointcloud$ (positions and normals) for the observed surfaces. The point cloud is not expected to capture every surface; entire sides of the input model may be missed. 
The images are acquired by walking around the model and taking photos, with enough coverage that at least one sheet plane per part is observed well enough to be partially reconstructed.
The point cloud is expected to have the model separated from background points as well as being oriented so that the object is approximately vertical, which in practice means the user indicates a ground plane and rough bounding volume for the object. 


\paragraph*{Part Identification.}
The goal of the first stage is to recover a set of parts, each with a rough approximation of $\pose$, $\depth$, $\shape$. These parts should closely match $\pointcloud$ and, although they may not strictly satisfy assembly constraints, they should maximise assembly feasibility in order to serve as a plausible basis for subsequent refinement.
Our approach is to initially detect an over-complete set of candidate parts by searching for planes that could be wooden sheets in the point cloud and extracting initial thickness and shape from point cloud features. We merge candidate parts when they are better represented as single sheets, e.g. when they originate from disparate points observed from opposite sides of the sheet. From among these candidates, we extract a subset $\parts$ of the parts with the best coverage of $\pointcloud$ that is also plausible from a fabrication standpoint; parts should not represent cuts through implausibly thick wood planes, and every part should subtend a minimum volume free of overlaps with other parts.

\paragraph*{Model Assembly}
Having decided on an initial set of parts $\parts$ from the part identification stage, we can proceed to infer the \textit{assembly}---defined by the connections between parts---and refine the part orientations to regularize the angles in the design. 
Detecting the correct joinery between parts is challenging since it is ``hidden'' beneath the surface geometry of the model; we address this problem with two key insights. First, we can identify a small set of types of connections possible with our fabrication assumptions, which significantly reduces the search space. Second, we can use image cues, such as the presence of seams or material changes visible in the wood, to identify regions where a connection interface is likely to exist (if geometric cues are insufficient). We use these ideas to disambiguate connections, followed by a global optimization step that aligns near-orthogonal connected parts while staying close to the point cloud.

\paragraph*{Geometry Refinement.}
The final step is to refine the geometry of the parts to obtain a fabricable model consisting of concise parametric curves, ensuring that the final model is consistent with the assembly constraints and represented using only the necessary number of primitives. We strive for simplicity to facilitate editing and because it is also usually consistent with how objects are designed.
We utilize the shapes visible in the input images to obtain more accurate cut path contours: Since the sheet plane position $\pose$ has already been identified for every part, we can use it as a reference to drive a multi-view image segmentation after projecting each image into this plane. We then apply a curve fitting algorithm over the resulting segmentation mask boundary which preserves the assembly constraints while globally minimizing an energy function that balances complexity and accuracy. Our final result is regularized by aligning curves and lines in the resulting shapes. In practice, the refined shapes may reveal new connections, so we iterate model assembly and refinement until no new connections are found.

\section{Technique}


\subsection{Parameters}
Given oriented point cloud $\pointcloud$, we define the global diameter $\diameter$ as the points' maximum bounding box dimension. Our method has many parameters which depend on the scale of the model, which is arbitrary; we therefore define these parameters in terms of $\diameter$. 
For detecting orthogonality and parallel features throughout our pipeline, we have a global angle threshold $\alpha$. 

\subsection{Part Identification}


In this stage, we both identify parts that comprise the model and estimate each part's rigid pose and rough shapes, as a basis for subsequent refinement. Our strategy for identifying the set of parts builds on the assumption that they are cut-outs from flat sheets and the fact that planes can be easily detected from 3D point clouds. Based on this insight, we can begin with primitive detection on the point cloud, 
followed by generating 3D parts from extrusions of paths in those planes, which amounts to detecting a cut path $\shape$ and the sheet thickness $\depth$. Note that using all detected planes as potential part planes results in many more parts than are actually in the model, as any given part has at least two planar surfaces, and more for straight line cuts; ultimately, only a subset of planes should be used. Our approach is to first generate part geometry for \textit{all} detected planes to form an over-complete set of candidate parts, then optimize for the subset that best approximates the model while representing a feasible construction.



\subsubsection{Primitive Detection and Adjacency}

We employ Efficient 
RANSAC~\cite{schnabel2007efficient} to segment the oriented point cloud into clusters of points that fit planes and cylinders; the points are roughly contiguous sets on those primitives.  The planes correspond to wood sheets, as well as any straight line cuts.  The points that fit better to cylinders tend to lie only on curved cut paths.  We used an inlier threshold of $\tau = D/300$ for the RANSAC algorithm.

We say that two primitives are adjacent if the minimum distance between their respective point sets is less than $\tau$.  We track adjacency between plane primitives and other primitives; let $\adjacency_i$ be the set of primitives adjacent to plane primitive $i$.  We later use adjacency to guide part depth estimation and to help bound the cut path for each part.

At this stage, every plane primitive now corresponds to a part $P_i$ with transformation $T_i$ that maps the $x$-$y$ plane to that primitive plane.  This set of parts is highly redundant; e.g., a fully observed cuboid part would have six planes corresponding to the sides of the part, and thus this one part would initially be over-represented by six parts.  We address part redundancy in the part selection phase at the end of this section.  First, we estimate the cut path $\Theta$ and thickness $d$ for every candidate part.

\subsubsection{Cut Region Approximation}
\label{sec:cutregion}
Each detected plane is a candidate to be a part's wood sheet plane.  We approximate its cut path as follows: collect the plane's associated points, transform them by $T^{-1}$ and project them onto the $x-y$ plane, convolve (in 2D) with a Gaussian kernel ($\sigma=D/400$), sample the resulting field over a regular grid, and extract a set of isocontours (isovalue of $1/\sqrt{e}$ \hle{where $e$ is Euler's number}). We assign the contour with the largest enclosed region, along with any countours inside that region, to be the approximate cut path. Note that interior contours enable parts to have holes cut into them. For more details, see the supplementary material.

\subsubsection{Initial Sheet Thickness Estimation}
\label{sec:thickness}


The thickness of a part can be determined in two ways: by measuring the cut surface or by the distance between its front and back sheet planes.
As we may not always see the back face (e.g., a part facing downward near the floor that happens to be missed during capture), we initially estimate the cut surface width to determine thickness. 

For each candidate part $\parti_i$, we estimate the cut surface width using a discrete plane sweep approach.
Specifically, we collect the points associated with the part plane's adjacent primitives $\adjacency_i$. 
We then move the plane of $\parti_i$ in the direction opposite its normal in discrete jumps $\Delta d$ and form a histogram $h(k)$ of adjacent points within $\Delta d/2$ of the plane offset by $k \Delta d$.  We set $\Delta d=\diameter/100$.

\begin{figure}
    \centering
    \includegraphics[width=0.7\textwidth]{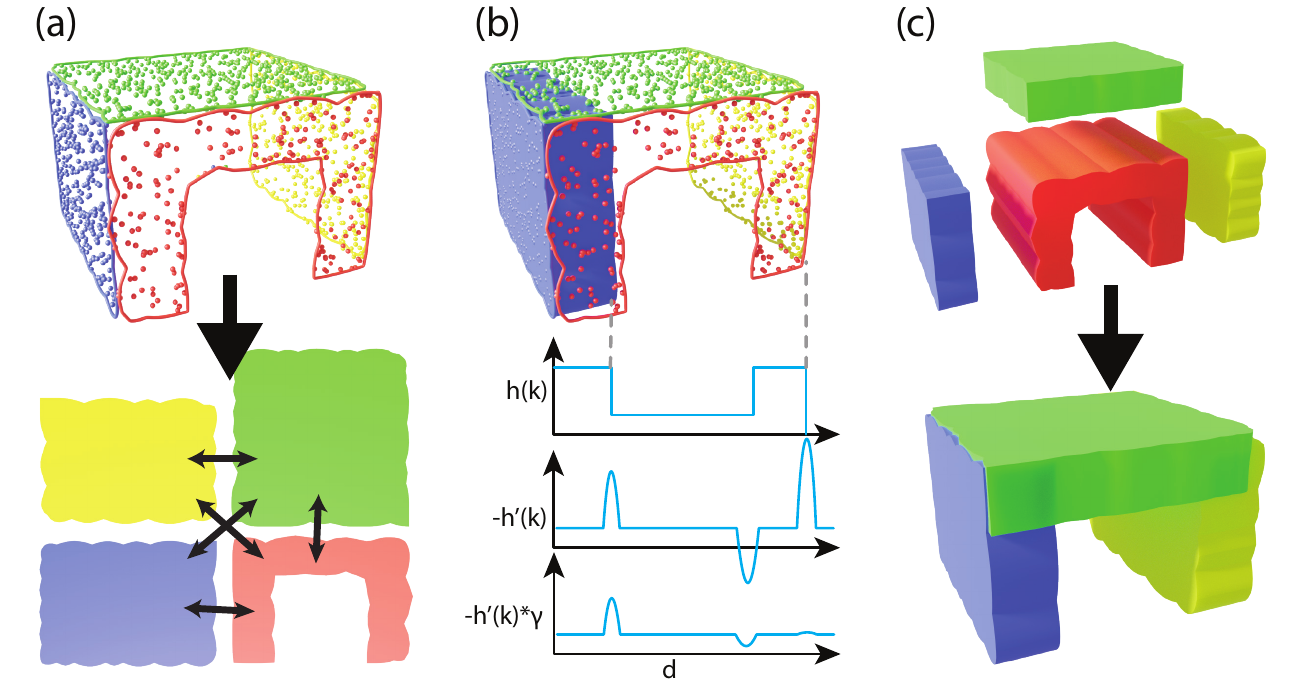}
    \caption{
    Overview of part identification. (a) We begin by detecting primitives and adjacencies between them, along with initial shapes for each plane. (In this example, the far side was not observed, thus the red curve does not have a symmetric counterpart). (b) We illustrate finding the thickness of the blue part by sweeping its plane and counting points on adjacent primitives near the sweep plane, forming histogram $h(k)$. We identify the plane-sweep thickness as the largest robustly identify peak, weighted by a spatial discount factor $\gamma(k)$. (c) shows the selected subset of generated candidate parts for the final model.
    }
    
    \label{fig:selectionoverview}
\end{figure}

When the plane sweeps past its cut surface, we expect a sharp discontinuity in the histogram, i.e., a peak in $-h'(k)$.  We compute $-h'(k)$ using finite differences and identify robust peaks with non-maximum suppression.  The largest peak may not correspond to the correct thickness; peaks closer to the part plane should take precedence (see Figure \ref{fig:selectionoverview} (b)). To address this, we weight the peak magnitudes with a spatial discounting factor, $\gamma(k)=\exp(-c_{\mathrm{falloff}}\cdot k \Delta d)$.  We set $c_{\mathrm{falloff}}=5/\diameter$.  The part thickness after this stage is set to $d = k_{\mathrm{peak}} \Delta d$, where $k_{\mathrm{peak}}$ is the bin of the largest robust, spatially-weighted peak. 

%


We additionally estimate thickness by considering planes that could be the opposite side of a part's wood sheet.  Given part $P_i$, for each part $P_j$ with opposite sheet normal, we transform its cut path $\Theta_j$ by $T_j^{-1}$ and project it onto the plane of $P_i$.  If  $\Theta_i$ and $\Theta_j$ overlap, we consider the distance between the planes of $P_i$ and $P_j$ to be a candidate thickness; we take the min over all of these thicknesses, call it $d_\mathrm{opposite}$.  If $d_\mathrm{opposite}$ is within $\Delta d$ of $d_i$ computed above, we then set $d_i = d_\mathrm{opposite}$.  During this step, we also record all other opposing parts $j$ with cut path overlap that are within $\Delta d$ of the final thickness (not just the closest part); call this set $O_i$, to be used later for merging parts.


\subsubsection{Part Selection}

The final step in the part identification stage is to select a subset of parts to be assembled and refined in the next stages.  We first reduce the number of parts through pruning and merging steps, and then perform a global optimization to give a set of parts that covers $\pointcloud$ well without too much overlap between parts.

To prune the part set, we first adopt a heuristic: parts are unlikely to be much deeper (thicker) than they are wide.  For example, if we have a cuboid part that is 30cm~x~30cm~x~1cm, we will prefer a 30cm~x~30cm face cut into a sheet 1cm thick over a 1cm~x~30cm face cut into sheet 30cm thick.  We prune as follows: if the estimated thickness of a $\parti_i$ yields a cut surface with total surface area greater than five times the area of $\shape_i$, we discard it. In Figure \ref{fig:selectionoverview}, the red part is one such candidate and is thus discarded.


Next, we merge part candidates if we have evidence they correspond to a single cut of the same wood sheet.  In particular, for part $P_i$, the set $O_i$ contains parts with opposite faces, cut paths overlapping $P_i$'s, and with planes roughly $d_i$ away from the plane of $P_i$.  These opposing parts are likely part of the same cut from the same sheet, and thus we transform and project the cut path $\Theta_j$ for each part $P_j \in O_i$ into the plane of $P_i$ and take its union with $\Theta_i$, after which we discard part $P_j$.  This step is useful to recover parts only reconstructed partially from different sides due to occlusions. We perform this process recursively until no such opposite-and-overlapping candidates remain. Note that, when merging $P_j$ into $P_i$, we also merge the adjacency sets, i.e., $\adjacency_i \leftarrow \adjacency_i \cup \adjacency_j$, useful later for geometry refinement.



Among the remaining parts, we generally still have over-representation, i.e., parts that overlap each other heavily.  Some amount of overlap is tolerable.  E.g., two cuboid parts that meet at a corner may overlap because it is unclear which part goes all the way to the corner and which has a cut face that abuts that other part; we will allow this small overlap and disambiguate it in the next section.  We now pose a (non-trivial) discrete optimization problem: select the set of parts $\parts$ that minimizes distance between $\pointcloud$ and $\parts$ without conflict. We say that a part is in \textit{conflict} if more than half of its volume overlaps with other parts in $\parts$.  We use simulated annealing to optimize for the final subset, where our energy function is the total squared distance between points in $\pointcloud$ and $\parts$.  We scale down all dimensions by $1/\diameter$ to ensure consistent behavior regardless of scale. To solve this problem, we employ the Metropolis-Hastings (MH) algorithm with transitions consisting of either adding or removing parts from the solution set, while prohibiting changes that lead to conflicts. Simply adding or removing individual parts at each step, however, results in poor convergence due to the large number of potential conflicts, so we additionally permit ``replacement'' transitions in which a part in the set may be swapped for another outside the set if adding the outside part would have otherwise caused it to be in conflict. To be precise, our proposal distribution is the result of the following decision process: with equal probability, either choose a part uniformly at random to add or remove from the set, or perform a replacement move between two parts as discussed above. We run MH for 1000 iterations with start and end temperatures of 10 and 0.1, respectively, and find that results typically converge within 10 seconds.

\subsection{Assembly Refinement}


Model assembly involves determining which pairs of parts are connected and the surfaces at which parts make contact, known as interfaces. Our approach is to first identify the connections for individual pairs of parts followed by a global alignment step that ensures manufacturability constraints over the graph of connections. The result of finding and classifying the connections between parts is a set of planar \textit{interface surfaces} along which the pairs of parts join, shown in blue in Figure \ref{fig:connectiontypes}.

 
\subsubsection{Connections}

\label{sec:connections}
\begin{figure}
    \centering
    \includegraphics[width=0.7\textwidth]{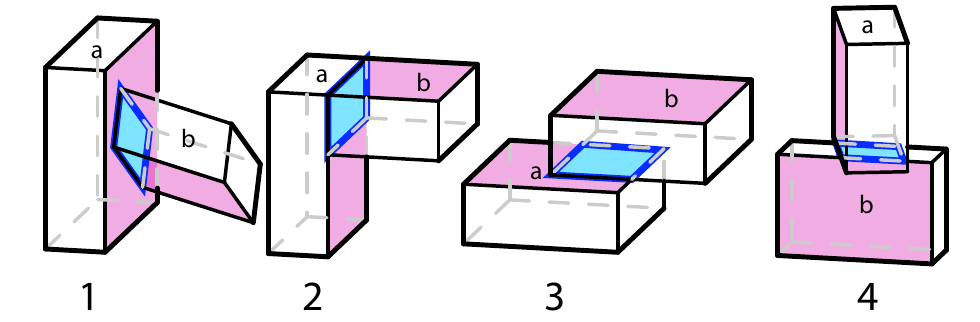}
    \caption{Types of unique connections based on which surfaces make contact. Connection interfaces are shown in blue. Each part's sheet plane is highlighted in pink. 
    (1) cut to center face
    ; (2) cut to corner face
    ; (3) face-only; (4) cut-only.
    In (1), we illustrate that the sheet planes need not be orthogonal, made possible with a bevel cut to b.}
    \label{fig:connectiontypes}
\end{figure}
\begin{figure}
    \centering
    \includegraphics[width=0.7\textwidth]{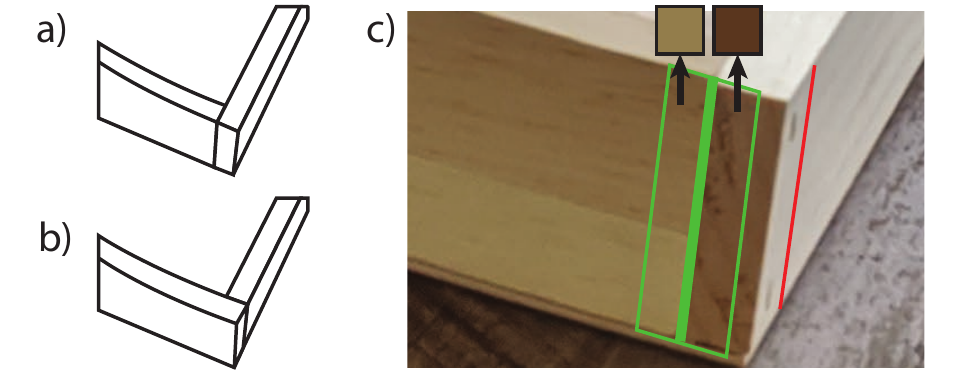}
    \caption{Two competing corner configurations (left), and the seams in the image used to disambiguate them. The color difference suggests that the green line is the boundary between parts, rather than the red line.}
    \label{fig:seams}
\end{figure}

First, we identify pairs of parts to connect.  Specifically, if the minimum distance between the surfaces of two parts is less than $\connectionthreshold = D/30$, then they are connected.

Next, we determine the type of connection between the parts. Based on our fabrication assumptions we identify 4 possible types of unique connections according to the types of surfaces that make contact, illustrated in Figure~\ref{fig:connectiontypes}. We exclusively deal with connection types 1 and 2, as we did not observe the other types in any of our example models. 
If the sheet planes of parts a and b are not orthogonal, as shown in Figure~\ref{fig:connectiontypes} (1), we use bevel cuts to satisfy the planar contact. To detect type 1 connections between two parts a and b, we determine whether a and b form a T-junction by checking if b terminates at a's sheet plane, and in the reverse case, if a terminates at b's sheet plane. If both a and b terminate at the other part's sheet plane, we say they meet at a corner and classify the connection as type 2. For more details on the geometry involved, see the supplementary material.




\subsubsection{Disambiguating corners}

A type 2 connection has two equally viable solutions for how the two parts meet.  To resolve this ambiguity, we look for evidence of a seam in the images that may indicate which part extends to the corner (see Figure \ref{fig:seams} (a) and (b)).  Given the approximate part geometry, we can determine which images have an unoccluded view of the (possible) seam.  For each such image $I_i$, we compute two measures of visual discontinuity.  First, we compute the gradient in the direction orthogonal to the seam at each pixel along the seam and average their magnitudes; call this value $g_i$.  We additionally compute a very coarse gradient across the seam by computing the average color within a rectangle of width $\tau_c$ on either side of the seam (see Figure \ref{fig:seams} (c)) and compute the magnitude of the difference of the colors, call it $\bar{g}_i$.  The seam score for this view is just $g_i + \bar{g}_i$.  We then average this score across all views of the seam to compute the final seam score and choose the type 2 configuration with the higher seam score.  If there are no views of a seam (e.g., if it is against the floor and thus not viewable), then we assign it a seam score of 0.01 (where pixel intensities range form 0.0 to 1.0) so that it can still be chosen if the other seam score is low (not a seam).  


\subsubsection{Interface surfaces \& constraints}
\label{sec:constraints}

After determining the connection types, we compute the finite interface surfaces within each interface plane indicating where the parts make contact. These give an estimate of where the final parts \textit{will} make contact for purposes of constraining their shapes. In both cases 1 and 2, we find this by intersecting the solid shape of part b with the abutting plane of part a, offset by $\connectionthreshold$ toward part b to correct for any gaps between parts caused by $\shapeinitial$. This interface can only take the form of one or more rectangles, each with width equal to the thickness of part b.

The constraints imposed by an interface surface on part b's cut shape are line segments that the shape cannot cross (without butting into another part), formed by the projection of the above surfaces onto the plane of part b. 
There are also additional constraints imposed by type 2 connections: In our final shape, the parts should meet perfectly at the corner determined by the line of intersection of the sheet planes from a and b on the outer surface of the corner. 
Finally, aside from connections, we have the constraints implied by adjacent plane primitives in $\adjacency_i$ that meet the part plane with a convex interior angle, since neighboring cut surfaces are evidence of the shape boundary. We exclude adjacent planes associated with connected parts, as satisfying the interface constraints should take precedence. These planes only correspond to planar cut faces; we do not include adjacent cylinder primitives, as we found their fits to be less faithful to the curves that they tend to fit. 
These \textit{constraint segments} in each part's sheet plane are used during geometry refinement so that the result conforms to detected surfaces and the precise connections inferred above. They define a half-space in the plane that belongs outside the cut region, for points that project to the line within the segment boundaries. For each constraint segment, we also make note of the angle of the interface surface relative to the part so that we can accurately define the bevel cut angle to enable this contact later.

We also expect the final model to lie flat against the ground. To incorporate this constraint, we add a ground plane, positioned at the lowest point of the input geometry, and form a fake "part" which we include in the above assembly analysis above to obtain additional contact constraints for any parts in contact with the floor.

\subsection{Geometry Refinement}
In the final stage of our pipeline, we refine the cut path for each part to be consistent with assembly constraints and image evidence and to be represented with concise, piecewise smooth curves. We do this in two stages: Image co-segmentation and constrained curve fitting.

\begin{figure}
    \centering
    \includegraphics[width=0.7\textwidth]{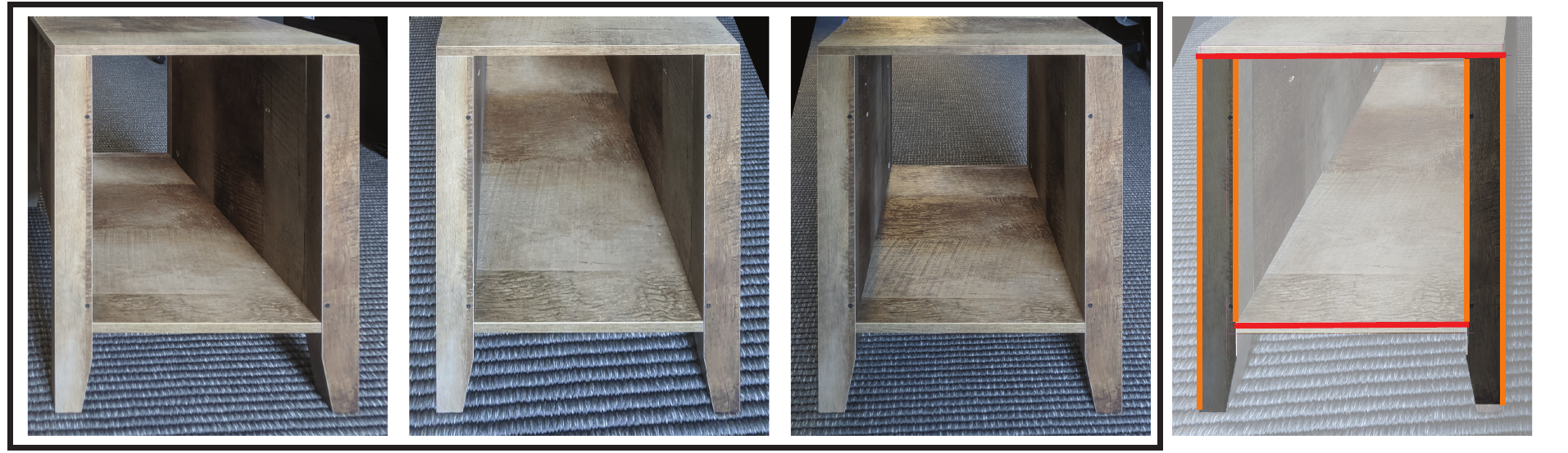}
    \caption{Several rectified images (left) used to obtain a segmentation mask of the cut region of a part (right). The regions where $\mask=1$ are shaded dark. 
    Interface constraints are shown as the red regions; other surface constraints, such as those arising from neighboring detected planes, are shown as orange lines.}
    \label{fig:segmentation}
\end{figure}

\subsubsection{Joint Image-Based Segmentation}


We leverage multiple views to optimize for a binary segmentation mask $\mask$ in the sheet plane representing the cut region for each part. Taking inspiration from~\cite{kowdle2012multiple}, aimed at segmentation and plane reconstruction, we pose our multi-view segmentation problem as an MRF optimization.  Unlike~\cite{kowdle2012multiple}, we use the known part plane as reference, leverage visibility cues in the rest of the reconstruction, and incorporate assembly constraints in the segmentation.

In particular, for part $P_i$, we project each image $I_j$ from its camera viewpoint onto the nearest part plane of $P_i$.  We resample the projection in the plane to form rectified image $\tilde{I}_j$. We then optimize for $\mask$ by defining a binary MRF on the set of pixels $\coordinates$ in $\mask$ with 4-connected grid edges denoted by $\edges$, with the energy

\begin{equation}
    E_{MRF}=\sum_{\mathbf{x}\in\coordinates} \!\! E_d(\mathbf{x}) + \lambda_s \!\!\!\!\! \sum_{(\mathbf{x}, \mathbf{y})\in\edges} \!\!\!\! E_s(\mathbf{x},\mathbf{y})
\end{equation}

\noindent
where $E_d$ is an appearance-based cost, and $E_s$ is a pairwise edge-sensitive smoothness term.  These energy terms depend on the per pixel mask labels $\mask(\mathbf{x})$ (1 for inside the cut path, 0 for outside), and we seek the lowest energy labeling with respect to all views together.  We set $\lambda_s = 50$.

\paragraph*{Data term}
We model appearance using Gaussian Mixture Models (GMMs) with 5 components for the colors (in LAB space) inside and outside the part's current cut region for each view, which gives us probabilities $P^1_j$ and $P^0_j$ that a pixel belongs inside or outside, respectively, for view $j$. For the interior, we erode the cut region by $\connectionthreshold$ (multiplied by the world-to-pixel scale factor) since the pixels near the boundary of the initial cut region are uncertain, 
and then consider only the subset of those pixels potentially visible to view $j$; for this purpose we construct a visibility mask $\occlusionmask_j(\mathbf{x})$ which is 0 if a ray from view $j$ to pixel $x$ intersects another part before reaching the part plane and 1 otherwise.  For the exterior, we similarly dilate the cut region and consider all pixels outside of it with $\occlusionmask_j(\mathbf{x}) = 1$. 


We now define the data term as:
\begin{equation}
E_d(\mathbf{x})= -\frac{1}{N_V  (\mathbf{x})}\cdot\sum_{j=1}^N \occlusionmask_j(\mathbf{x})\cdot\log(P^{\mask(\mathbf{x})}_j(\mathbf{x}))
\end{equation}
where $N_V(\mathbf{x})=\sum_{j=1}^N\occlusionmask_j(\mathbf{x})$ is the number of views that can see pixel $x$ on the part plane.   If $N_V$ is 0, we set $E_d$ to 0 regardless of $\mask(\mathbf{x})$.  In general, some ``outside'' pixels may belong to other parts with similar appearance, so we modify $E_d$ to incorporate the constraint segments computed earlier: $E_d(\mathbf{x})$ is set to $\infty$ for $\mask(\mathbf{x})=1$ if $\mathbf{x}$ is in the excluded region of any of the constraint segments. 
The result of using these constraint segments for segmentation is shown in Figure \ref{fig:segmentation} (b). The red interface constraints prevent the mask region from including surfaces from adjacent parts, which purely appearance-based segmentation would do.



\paragraph*{Smoothness term}
We define $E_s$ using a contrast-sensitive Potts model to regularize the result while aligning the mask with high-contrast regions which we expect at shape boundaries.

\begin{equation}
    E_s(\mathbf{x}, \mathbf{y})=|\mask(\mathbf{x}) -\mask(\mathbf{y})|\exp\left(-\frac{1}{N_V(\mathbf{x})\sigma_s^2}\sum_{j=1}^N \occlusionmask_j(\mathbf{x})(\rectifiedimage_{j}(\mathbf{x})-\rectifiedimage_{j}(\mathbf{y}))^2\right)
\end{equation}
where $\sigma_s$ (set to 50) controls the strength of the smoothing penalty falloff as contrast increases. Analogous to how we modify $E_d$, we also set $E_s(\mathbf{x}, \mathbf{y})$ to zero in cases where the 3D locations corresponding to $\mathbf{x}$ and $\mathbf{y}$ straddle a constraint segment, 
to encourage the boundary of $\mask$ to adhere to these known edges; i.e., there is no penalty for label change at these boundaries where label changes are likely.

In practice, for efficiency, we only consider views $I_j$ for which $V_j(\mathbf{x})=1$ for at least half the pixels inside $\Theta$, and then use the top seven views sorted by how close their central viewing rays are aligned with the plane normal.  Note that this set of views may come from one or both sides of the part. 
We solve the MRF using graph cuts ~\parencite{boykov2001interactive} to obtain the final mask. We also re-use the resulting $\mask$ (cut path) to learn more accurate GMM parameters, and re-run the above algorithm once more to slightly improve results. 

\paragraph*{Updating model topology}
It is possible for $\mask$ to have more than one connected component after optimizing $\mask$ with the assembly constraints, if multiple parts were detected as one in previous steps (as is the case in Figure \ref{fig:segmentation}; the legs are forced into separate pieces by the assembly constraints). We restructure the model in these cases by adding each \hle{connected component in $\mask$} as a separate part. Finally, \hle{we rerun the assembly stage to find new connections and constraints due to the updated shapes and potentially separated parts. We repeat the segmentation and assembly steps until no new connections are found} (we observe at most 1 or 2 iterations in our experiments).

\subsubsection{Global Alignment}

Before extracting a final CAD model, we align connected parts that are close to orthogonal, as right angles are a feature of many manmade designs. This is important because it simplifies the fabrication process considerably as well; parts connected at right angles only require orthogonal cuts, which can be made with a wider variety of tools. Since this optimization only concerns small perturbations to the orientation, we represent each part $\parti_i$ by its sheet plane $\plane_i$ 
and optimize over plane parameters (normal $\mathbf{n}_i$ and offset $o_i$) such that detected orthogonal connected parts are aligned. We minimize the total squared distance of the planes to their detected point sets to regularize the result. We take an approach similar to~\cite{li2011globfit} for plane alignment; we find
\begin{equation}
\min_{\forall\mathbf{n},o}\sum_i E_p(\pointcloud_i, \plane_i) 
\end{equation}
subject to $\mathbf{n}_i\cdot\mathbf{n}_j=0$ 
for all $i,j$ for which $\parti_i$ and $\parti_j$ are connected and the angle between $\mathbf{n}_i$ $\mathbf{n}_j$ is within $\alpha$ of $90^\circ$, where $\pointcloud_i$ is plane $i$'s point set, 
and $E_p$ is the total squared distance. To ensure unit normals, we represent each $\mathbf{n}_i$ using 2 angle parameters.
We solve this global optimization problem using a sequential least-squares quadratic programming (SLSQP) solver, and then update $\pose_i$ to align the parts with these new plane parameters. 

\subsubsection{Thickness Regularization}
Typically, an object is constructed by cutting from a small number of wood sheets, with a small number of thicknesses.  However, since our initial thicknesses are based on analysis of noisy point clouds (\ref{sec:thickness}), we typically estimate a different thickness for every part.  Minimizing the number of distinct thicknesses in a model makes it more practical to build and therefore more plausible. Thus, we cluster thicknesses by averaging any part thicknesses that differ by less than a threshold $\connectionthreshold$. 

\subsubsection{Constrained Curve Fitting}
Our curve fitting approach draws ideas from from prior work that trades-off global fit accuracy with curve complexity~\cite{plass1983, farin2002curves, fleishman2005robust} as well as prior work that favors straight edges and sharp corners~\textcite{dominici2020polyfit}, and applies them to the context of handling imperfect binary masks, with certain known edges that the solution must adhere to. 

For each part, the output of the segmentation step is a cut path defined by the raster boundary of a segmentation mask; it is neither exact nor concise.  Our final step is to extract a CAD representation of this path by fitting a low-dimensional 2D shape representation that approximates the segmentation boundary while adhering to any contact constraints. 
In related works on vectorization, the perceptual criteria of accuracy and simplicity, along with continuity and regularity, are prominent objectives (~\cite{hoshyarivectorization,Kopf2011,dominici2020polyfit}).  We find these objectives to be well-suited to our problem: We desire a shape that is close to the input boundary while adhering exactly to the contact constraints, and which also provides a simple, continuous explanation for the input mask boundary, while capturing regularity in the man-made objects that are our focus.



We represent cut paths as closed $G_0$ continuous polycurves consisting of connected cubic Bézier curves and straight line segments, where we call the endpoints between neighboring segments \textit{nodes}. 
For each part, our algorithm takes as input the raster boundary of the segmentation mask, a (clockwise) ordered set of 2D points $X$.  We restrict nodes to lie on points in $X$ and therefore have a discrete set of possible nodes, where each segment is the least-squares best fit for the range of data points between nodes. 

We solve for the curves by building on the dynamic programming approach outlined in~\cite{plass1983}.  Let $e_{ij}$ be the cost of fitting a curve to the subrange between $X_i$ and $X_j$, which is the sum of squared point distance error plus a constant curve cost $c_1$. 
We define a sub-total energy $E_{ij}$ as the least total error over all possible choices of nodes between $X_i$ and $X_j$, giving rise to the recurrence relation $E_{ij}=\min_k(E_{ik}+e_{kj}), i<k<j$, allowing us to solve for the optimal node locations using dynamic programming. Support for $G^1$ continuity at curve transitions is added by pre-computing tangents at each point in $X$ using curves fit to a local point neighborhoods, and constraining the end tangent directions of curves during fitting. 


We add support for different curve types by extending $E_{ij}$ to $E_{ijk}$, where $k$ is the type of the curve ending at $X_j$; sub-total energies are now computed by summing over all previous curve types, as well as all previous nodes. We now have separate curve costs $c_k$ for each type. We let $k=0$ indicate line segments, and set the cost $c_0=c_1/2$ to encode a preference for straight lines. For more implementation details, we refer the reader to the supplementary material. 
We also wish to capture both sharp corners and smooth transitions in our solution. Because the input may contain artifacts, and furthermore is not representative of the final boundary that adheres to all the desired constraints, we do not detect sharp corners in the input as is usually done in vectorization, but rather incorporate the choice into our curve fitting algorithm to encourage sharp corners that lead to a better fit to the data. 
To allow sharp and smooth tangent behavior at nodes, neighboring curves must be able to agree on either $G_1$ continuous or unconstrained tangents. To make this possible, we parameterize right end tangent behavior using two types $k=1$ and $k=2$, where each type is a cubic B\'{e}zier curve with a constrained and free right end tangent, respectively. This way, we can ensure that each curve's left tangent behavior matches the previous curve's right tangent behavior when computing $E_{ijk}$. Finally, we can filter out sharp corners by requiring that a curve with type $k=2$ must meet the next curve with an angle greater than $\minangle$. Because an unconstrained curve will always fit with smaller MSE error, cases where the tangent angle is $>\minangle$ represent unconstrained curves that differ significantly, and therefore should be preferred in the interest of accuracy. We set the curve cost $c_1 = (\frac{D}{1000W})^2$ where $W$ is the width of the input boundary's mask, and $\minangle=10^\circ$ in our experiments.

\begin{figure}
    \centering
    \includegraphics[width=0.75\textwidth]{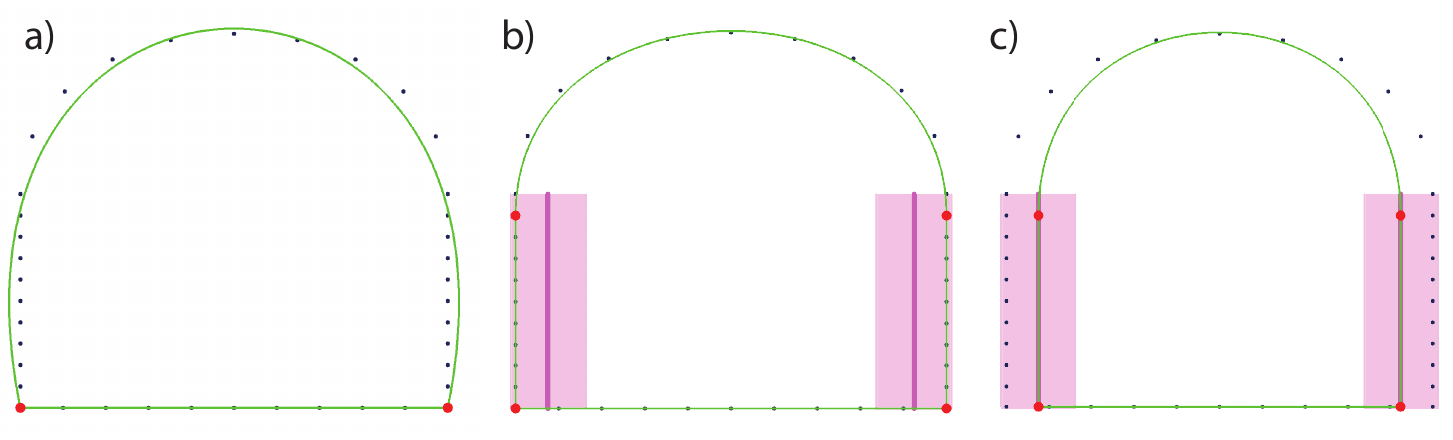}
    \caption{Left to right: Polycurve fitting without and with constraints. Input points are shown in black, output curves in green. Constraints are shown as purple line segments; the pink region represents the tolerance of the constraint. In (a), the unconstrained solution fits a single curve to the upper points, while in (b) the constrained solution adds line segments which preserve smooth transitions, allowing them to be trivially displaced to adhere to the constraints while preserving smoothness (c).}
    \label{fig:curvefit}
\end{figure}

The constraint segments used in the segmentation stage 
are also used in curve fitting; we wish for the curve to "snap" to these segments wherever they are near enough, or if it would result in a simpler solution. 
We incorporate these constraints into our curve fitting algorithm by first identifying points in $X$ within $\connectionthreshold$ of a constraint segment, and forcing any segment fit to a range containing these points to be a straight line segment. 
The result is shown in Figure \ref{fig:curvefit} (a) and (b); the dynamic programming fit is guaranteed to produce line segments where they are needed, and does so while still fulfilling its other objectives of continuity and simplicity.

\paragraph*{Post-processing}
Having guaranteed straight edge segments in the \textit{vicinity} of constraint segments, we project the nodes bordering lines near these constraint segments to the exact lines of these constraints to obtain a fabricable solution (Figure~\ref{fig:curvefit} (c)). Neighboring curves are modified so as to preserve their tangent angle with the displaced lines. 
It is not always possible to ensure consistent tangent behavior in the above framework; transitions between curves and line segments are troublesome since the latter lack the degrees of freedom to adhere to the pre-computed tangents used for smooth transitions. The inherent order of curves considered by the dynamic programming algorithm prevents curves from correcting for the behavior of subsequent neighbors. 
We therefore apply an additional smoothing step in which corners below angle $\minangle$ are made smooth by altering curve tangents. This step is only done if the resulting change to the shape is not too drastic; we approximate this change by the total displacement of Bézier control points, which we limit to $\connectionthreshold$.

We additionally find lines that are parallel/orthogonal (within angle $\alpha$) and further align them. Constraint line orientations are left unchanged, and any edges nearly parallel to them copy their orientation, and likewise for nearly orthogonal edges. In many cases, 
this produces 90 degree angles in shapes where parts connect.

\section{Experimental Results}
\label{sec:eval}


\newcommand{\STAB}[1]{\begin{tabular}{@{}c@{}}#1\end{tabular}}


\begin{figure*}
    \centering
    \includegraphics[width=170mm]{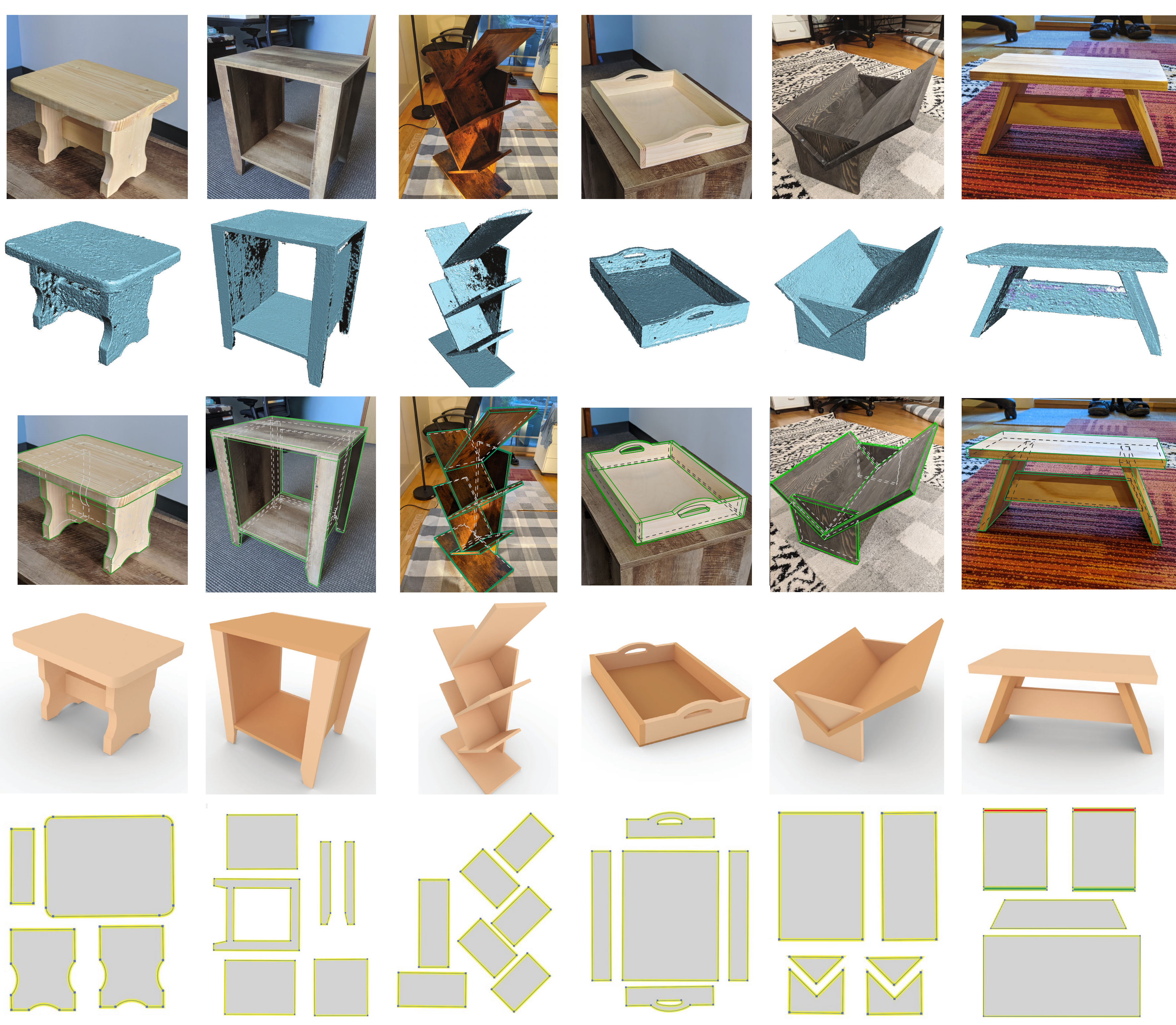}
    \caption{Results from six example models. Left to right: stool, nightstand, bookshelf, tray, bookholder, tilted stool. The results shown, from top to bottom: representative input image; reconstructed point cloud; our result superimposed on the input image; a clean render of the result; and the 2D cut paths, with curve endpoints shown as blue dots. Bevel cut surfaces are displayed in red and green, indicating that the surface normal faces into or out of the page, respectively.}
    \label{fig:resulttable}
\end{figure*}


We tested our algorithm on seven carpentered objects of varying complexity, i.e., varying numbers of parts with some objects exhibiting more difficult features, such as non-axis aligned parts (the diagonal bookshelf and bookholder), shapes with holes and curves (the stool and tray), non-orthogonal connections, and one which deviates slightly from our model assumptions in the form of smoothed corners and grooves in the sheet plane, as well as having some highly occluded part sheet planes (complex stool). We obtained fabricable reconstructions for six of the objects, and discuss the seventh as a limitation in the final section. 

\paragraph*{Experimental Setup}
We photographed our objects with a hand-held Google Pixel 3 camera in two distinct, well-lit indoor locations with a variety of backgrounds (different rugs, etc.). For each model, we took between 30 and 70 photos from viewpoints facing the object and situated approximately on a hemisphere around it. We used RealityCapture~\cite{capturingreality} to recover camera poses and semi-dense point cloud reconstructions which requires some minor user input to select the reconstruction region to isolate the object one desires to capture, in particular to omit the ground plane.

\paragraph*{Qualitative assessment}
Figure \ref{fig:resulttable} shows results for six of the models. In all cases, there were faces of the model that were either missing or incompletely represented in the point cloud (second row), such as the unseen undersides of the top of the stool and nightstand, as well as faces that are less well-textured or in shadow, as with some parts of the bookshelf. For all six models shown, we were able to generate fabricable results, shown in the fourth row. 
In the third row, we superimpose the reconstructed CAD model on the input image showing how faithfully our reconstructions fit to the observed object. 
Some minor failure regions include the horizontal pieces of the nightstand, which are slightly thicker and, in the bookshelf, two of the parts are slightly translated from their original position. In the final row, the cut shapes for all the parts in the output are shown as combinations of line segments and B\'{e}zier curves. The nodes, shown in blue, indicate the transitions between curves and may either be smooth or sharp corners. For the most part, these simplified curves are consistent with the input, but in the case of the tray, the smooth transitions on either side of the rounded handles, as well as on the bottom of the handholds, are sharpened. The cut shapes also reveal that the back side of the nightstand was detected as a single piece. In fact, that piece is made up of 4 smaller pieces, as shown in Figure \ref{fig:limitations} (a). Though we \hle{do not} decompose detected parts based on detected seams, this result can be ``fixed'' by adding cuts after the fact. \hle{To evaluate the importance of the methods in our technique, we also show some results with various parts of our pipeline simplified in Section 4 of the supplemental material.}

\paragraph*{Quantitative evaluation}\label{sec:quanteval}

To measure the accuracy of the final fabricable, simple CAD model to the original object, we use the RMSE distance of the point cloud to the model surface, multiplied by $1/\diameter$ for scale independence. As shown in Table \ref{tab:allresults} for the same five models in Fig~\ref{fig:resulttable}, this error is on the order of 0.4\% of the model width, indicating that in terms of geometric displacement, our reconstructed shapes remain true to the original objects. 


We also measure the number of incorrect connections, i.e. extra, missing, or misclassified connections. Among the five models, only the nightstand misses some connections; the four parts making up the back surface are detected as one part (see Figure \ref{fig:limitations} (a)). 

Another important measure of accuracy is in the representation of the cut paths themselves; smooth curves and sharp corners should be reflected in the final result. Two of our models contain smooth curves: In the case of the stool, these features were handled without problems; for the tray, some additional corners are present in the handle holes, as well as on either side of the arches.

\newcolumntype{M}[1]{>{\centering\arraybackslash}m{#1}}

\setlength{\tabcolsep}{-3pt}
\begin{table}
\small
    \centering
    \begin{tabular}{m{2.0cm}|M{1cm} M{0.8cm} M{0.8cm} M{1.2cm} M{1cm} M{1.3cm} M{1.3cm} M{0.8cm}}
    model 
    & \STAB{\rotatebox[origin=c]{60}{\#images}} 
    & \STAB{\rotatebox[origin=c]{60}{\#parts (actual)}} 
    & \STAB{\rotatebox[origin=c]{60}{\#parts (found)}} 
    & \STAB{\rotatebox[origin=c]{60}{RMSE (\% of $\diameter$)}} 
    & \STAB{\rotatebox[origin=c]{60}{identification}} 
    & \STAB{\rotatebox[origin=c]{60}{assembly}} 
    & \STAB{\rotatebox[origin=c]{60}{segmentation}} 
    & \STAB{\rotatebox[origin=c]{60}{curve fitting}} \\
    \hline
    stool & 68 & 4 & 4 & $0.433$ & 3s & 0.3s & 267s & 129s \\
    nightstand & 37 & 10 & 7 & $0.477$ & 25s & 25s & 498s & 70s \\
    bookshelf & 37 & 7 & 7 & $0.204$ & 8.7s & 4.7s & 507s & 31s \\
    tray & 30 & 5 & 5 & $0.259$ & 2.5s & 26s & 54s & 23s \\
    bookholder & 42 & 6 & 6 & $0.406$ & 5.2s & 27s & 131s & 23s \\
    tilted stool & 48 & 4 & 4 & $0.432$ & 15s & .038s & 199s & 20s \\
    \end{tabular}
    \caption{Statistics for five input models, showing the complexity of each model, the geometric accuracy of the reconstruction, and the time taken in each stage of the algorithm }
    \label{tab:allresults}
\end{table}

\section{Limitations and Future Work}
\label{sec:concl}

\begin{figure}
    \centering
    \includegraphics[width=80mm]{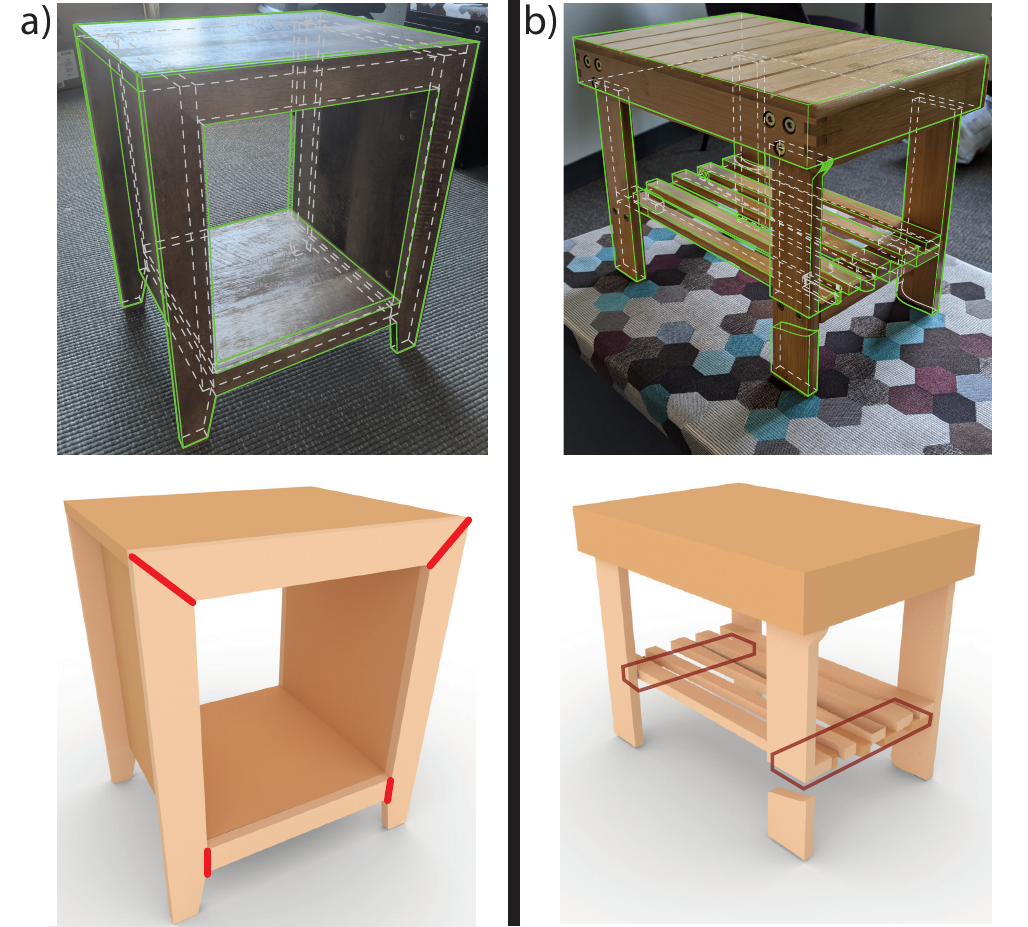}
    \caption{Limitations of our ability to capture part some part structures. In (a), the nightstand contains four co-planar parts that are merged into one in our result; the seams where they should be cut apart are highlighted in red. For the complex stool in (b), failure to detect a complete set of part candidates causes the result to be invalid.}
    \label{fig:limitations}
\end{figure}

%






A key feature and a notable limitation of our method is its reliance on image evidence.  The images enable the MVS reconstruction, corner assembly disambiguation, and recovery of nice cut paths.  However, we can only reconstruct what we observe sufficiently.  If a model has structure that hides some parts from view, it can be difficult or impossible for our method to accurately reconstruct the model. 
In the model in Figure~\ref{fig:limitations}b (``complex stool''), the parts highlighted in red are occluded by the boards directly above them, causing them to be detected as disjoint, floating pieces; the result is not fabricable or even connected. This might have been addressed by observing the underside of the red parts (and straightforwardly handling type 3 connections), but the underside was out of view.
In future work, it would be interesting to explore the use of additional priors or learned semantics to reconstruct objects with incomplete observations. For instance, if a model \hle{could not} be assembled only with detected parts (perhaps due to floating, unconnected pieces), we could potentially hallucinate new structures. Another way to address the method's reliance on image evidence would be to extend the system to enable real-time capture by incorporating new observations incrementally, and guiding the user to capture new images that resolve ambiguities or missing geometry in the model.

Our method has a number of thresholds, cost terms, and other parameters that affect the final quality of the result. Many of these parameters were related to object dimensions, to limit dependency on object size. 
The method, particularly the segmentation, and curve fitting stages, are sensitive to the choice of some of these parameters. We determined good values by experimentation up front and then used the same parameter values (or proportionality constants for parameters related to object scale) for every model. In the future, it would be desirable to to detect some of these parameters adaptively to improve the robustness of our method. \hle{ For instance, some specific stock thicknesses are more common as building materials, so part thicknesses could be used to detect the true object scale. This can further be used to ensure that parts can be made from readily available materials.}

It might also be worth exploiting alternative or complementary features derived from the input data. For example, we could use all seams in the wood as evidence of cuts to find connected co-planar parts as in Figure \ref{fig:limitations}a. To incorporate arbitrary seams robustly, we would need to account for other texture features, such as ever-present wood grain, that could be mistaken for cuts. 
In addition, man-made objects often have symmetries and repetitions, which could be exploited both to aid in detection and to further regularize the model. 
Lastly, in addition to planar surfaces detected from point clouds, we could also use the detected curved (cut) surfaces to guide segmentation. This would also help minimize the bias towards straight lines observed in the tray example.

\begin{figure}
    \centering
    \includegraphics[width=0.7\textwidth]{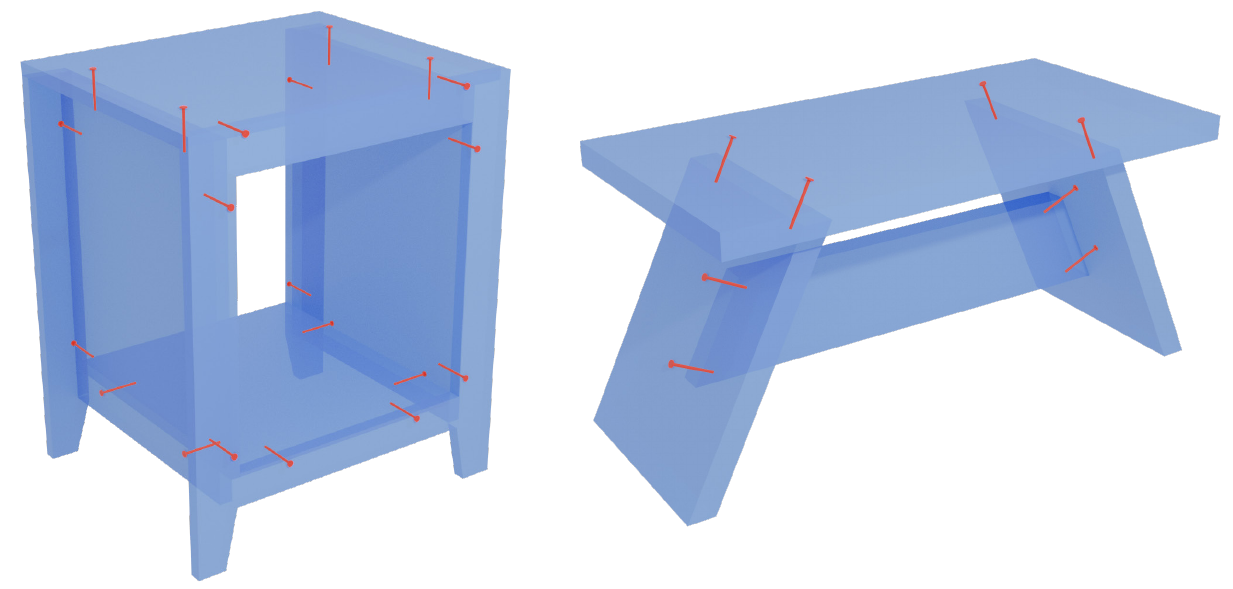}
    \caption{Examples of \hle{procedurally} generated connectors used for assembly.}
    \label{fig:nails}
\end{figure}


Since our method identifies where and how parts are connected by classifying the types of connection (Figure~\ref{fig:connectiontypes}), we can procedurally define connectors over the contact surface for each connection type. In our implementation we use \hle{two nails for each contact between parts, one at each end of the participating cut surface, as shown in} Figure~\ref{fig:nails}. This information can be used to create physical reproductions of the models, such as that shown in Figure~\ref{fig:teaser}. We leave it to future work to consider the problem of assembly in greater detail, in particular ensuring that connectors do not obstruct each other and determining easy-to-follow assembly instructions, which \hle{should ensure that a sequence of steps exists such that the model can be assembled without parts obstructing other parts, or the tools needed to assemble them}. One direction is to adapt prior work in automatic generation of assembly instructions to the output of our method~\cite{agrawala2003designing}.

Finally, it would be interesting to extend our method to more fabrication operations and objectives. 
Within carpentry, supporting more complex joinery would expand the possible geometries that could be reverse engineered. \hle{We currently assume planar contacts, both because they are common and because the internal structure of joints cannot always be observed without disassembly (such as with mortise and tenon joints). In some cases, however, it may be possible to parse joinery directly through image analysis. Optimizing joinery with respect to structural stability, similar to} \parencite{yao2017interactive}\hle{, or additional objectives such as fabrication cost and packing efficiency} \parencite{wang2021stateoftheart} \hle{could also guide reconstruction of joinery and other hidden structure}.

\section{Conclusion}
In this work, we propose a method for recovering accurate representations of built, carpentered objects from a set of photographs by working within the space of the fabrication process itself. This representation is both highly expressive and subject to real world constraints, as the process describes models that can be physically realized, making it a good candidate for solving inverse problems in 3D reconstruction.  Given enough images covering the surfaces, our solution in the carpentry domain can recover the parts and connections that comprise captured real world objects, complete with the simplified contours that most concisely describe the cut paths of the model, making it easy to edit with CAD software to create design variations. We hope this result will inspire future work at the intersection of fabrication and computer vision, leading to more end-to-end systems for 3D reconstruction that can take into account multiple materials and fabrication processes.



\section*{Acknowledgements}
This work was supported by National Science Foundation grants CCF-2017927 and EEC-2035717, UW Reality Lab funding from Facebook, Google, and Futurewei.  A. Schulz acknowledges the generous support of the Google Faculty Research Award.

\appendix
\section{Supplementary Material}

\subsection{Cut Region Approximation}
The function from which we extract isocontours to initially approximate the cut shape in Section 4.2.2 is
\begin{equation}
    f(\mathbf{x})=\sum_{\mathbf{q}} \exp(-\|\mathbf{q}-\mathbf{x}\|^2/(2\sigma))
\end{equation}

where $\mathbf{q}$ are the 2D points, and \(\sigma\) is ($\diameter$/400). We extract a level set $\{x\vert f(x)=1/e\}$ using the Marching Squares algorithm with a grid length of $4\sigma$.

\subsection{Inferring Connection Types}

\begin{figure}[b]
    \centering
    \includegraphics[width=0.45\textwidth]{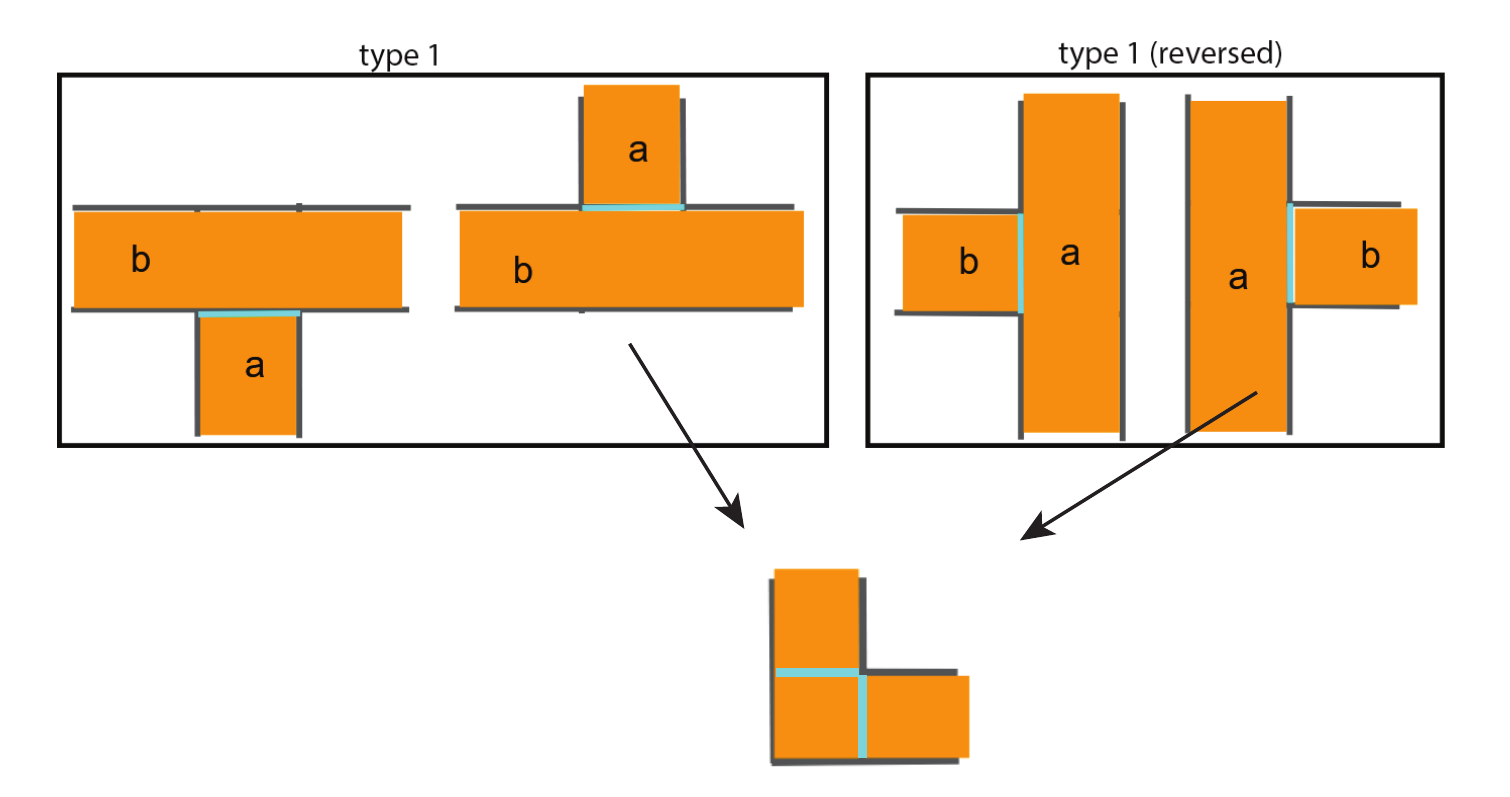}
    \caption{Top: Four cases of connection type 1 for a pair of parts, viewed from the side so that both parts' sheet normals are parallel to the page. Bottom: one corner configuration determined from the orientations of both candidate type 1 connections. Interface surfaces are shown in cyan.}
    \label{fig:cases}
\end{figure}

To detect whether part b approximately terminates at part a's sheet plane, thus forming a T-junction necessary for considering type 1 connections, we use part a's sheet normal $\mathbf{n}_a$, and a's sheet plane offsets $o_{\mathrm{min}}$ and $o_{\mathrm{max}}$ to compute the projected offsets of all points in part b: $z_{\mathrm{min}}=\min_{\mathbf{p}\in\parti_b}(\mathbf{p}\cdot\mathbf{n}_a)$ and $z_{\mathrm{max}}=\max_{\mathbf{p}\in\parti_b}(\mathbf{p}\cdot\mathbf{n}_a)$. If 
\begin{equation}
    z_{\mathrm{min}}<o_{\mathrm{min}}-\penetrationthreshold\label{eq:cond1}
\end{equation}
and
\begin{equation}
    z_{\mathrm{max}}>o_{\mathrm{max}}+\penetrationthreshold\label{eq:cond2}
\end{equation}
are both true, part b is not confined to either side of part a's sheet, preventing connection type 1. As long as only one is true, the connection is allowed; the interface plane offset is $o_{\mathrm{min}}$ if (\ref{eq:cond1}) holds, and $o_{\mathrm{max}}$ if (\ref{eq:cond2}) holds. Type 2 connections occur when a type 1 connection is valid for both orderings of parts a and b.

Consequently, there are actually 4 discrete configurations for a type 1 connection between two parts, as shown in Figure \ref{fig:cases}: For the connection and its reverse (where a and b are swapped), the connection may involve contact with one of two sides of the sheet plane (whether the interface plane offset is $o_{\mathrm{min}}$ or $o_{\mathrm{max}}$). For type 2 connections, the positions of these contacts for both the the type 1 connection and its reverse are used to determine the corner configuration, as shown in the bottom of Figure \ref{fig:cases}; knowing where the potential contact surfaces lie is crucial to knowing where in the images to look for seams.

We assume corners (type 2 connections) are right angles. Though it would not be difficult to allow them to vary, our choice of two "natural" corner configurations requiring only orthogonal cuts no longer makes sense; bevel cuts will be needed no matter what, so potentially more complex joints would need to be detected. Furthermore, non-orthogonal corners are uncommon.


\subsection{Curve Fitting}
The dynamic programming curve fitting algorithm can infer the optimal set of nodes from the input point set, with the caveat that it requires a starting point from which the optimal sub-ranges belonging to separate curves are determined. This starting point is necessarily a node, since it is the start of the first such range returned by the algorithm. A first instinct might be to choose a starting point that looks like it should be a corner; however, if the input shape has no true sharp corners, the exact tangent behavior at the start of the loop will depend on the behavior at the end of the loop, which violates the sequential order in which we find the curves.

Instead, we do the opposite: We look for a starting node in a region that is as flat as possible (which we determine using the curvature of a B\'{e}zier curve fit to a neighborhood of points centered at the query point). Such a region can be assumed to always exist for well-behaved inputs approximating continuous shapes, and allows the tangents at the loop boundaries to be assumed to be smooth. The downside is that the starting node often bisects a region that could be better described with a single curve or line segment. We therefore filter out the extra node whenever possible by merging collinear line segments (the most likely case).

In practice, it is very inefficient to consider every possible sub-range of points as a candidate curve. The space and time complexity of the dynamic programming algorithm is quadratic in the number of candidate nodes, and this number is potentially very large when the input is a dense bitmap mask boundary. So given a maximum number of candidate nodes $K$, we find the $K$ input points with the highest curvature and mark them as candidates, since such corners are likely to mark the boundaries between separate curves.

The full recurrence relation for our energy function $E_{ijk}$ is
\begin{equation}
    E_{ijk}=\min_{j',k'}(E_{ij'k'}+e_{j'jk'k}), i<j'<j
\end{equation}

where $i$ and $j$ are the start and end points of the considered range of points, and $k$ is the type of curve fit to the range ending at $j$ (so $E_{ijk}$ is the energy of a sequence of curves whose last curve has type $k$). Note that the sub-range energy $e_{j'jk'k}$ depends not only on the starting and end node indices, but also the \textit{type} of the previous and current curve. Because $k$ also defines whether a B\'{e}zier curve should use the fixed (precomputed) tangent at its last endpoint, this allows us to define the rules, detailed in Section 4.4.4, governing the behavior of neighboring curves, including the angles between their tangents where they meet.

\section{Evaluating Design Choices}


\hle{To gauge the necessity of some of the stages of our pipeline, we discuss how disabling or simplifying them impacts the quality of results.} 


\begin{figure}
    \centering
    \includegraphics[width=0.45\textwidth]{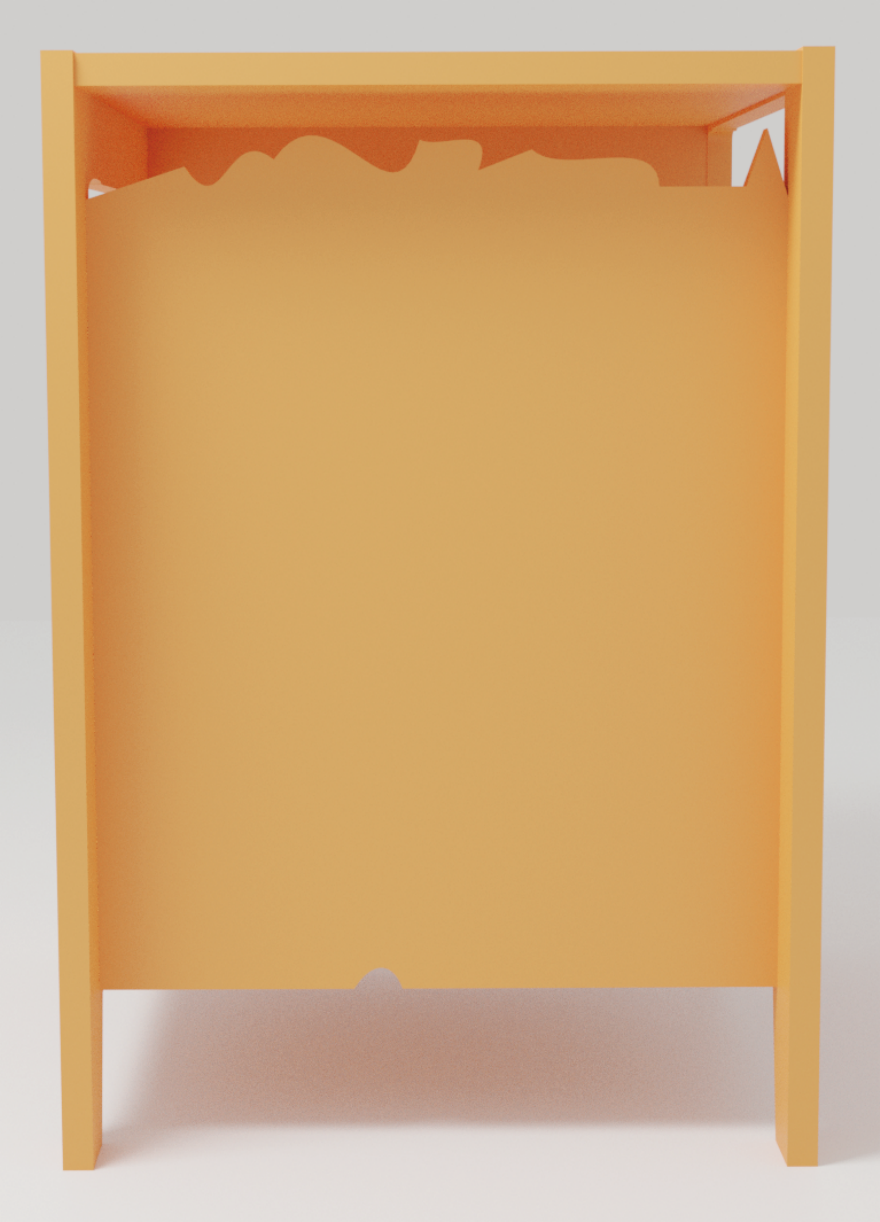}
    \caption{Without the image segmentation step, deficiencies in the input point cloud persist in the final result.}
    \label{fig:nosegmentation}
\end{figure}

\begin{figure}
    \centering
    \includegraphics[width=0.7\textwidth]{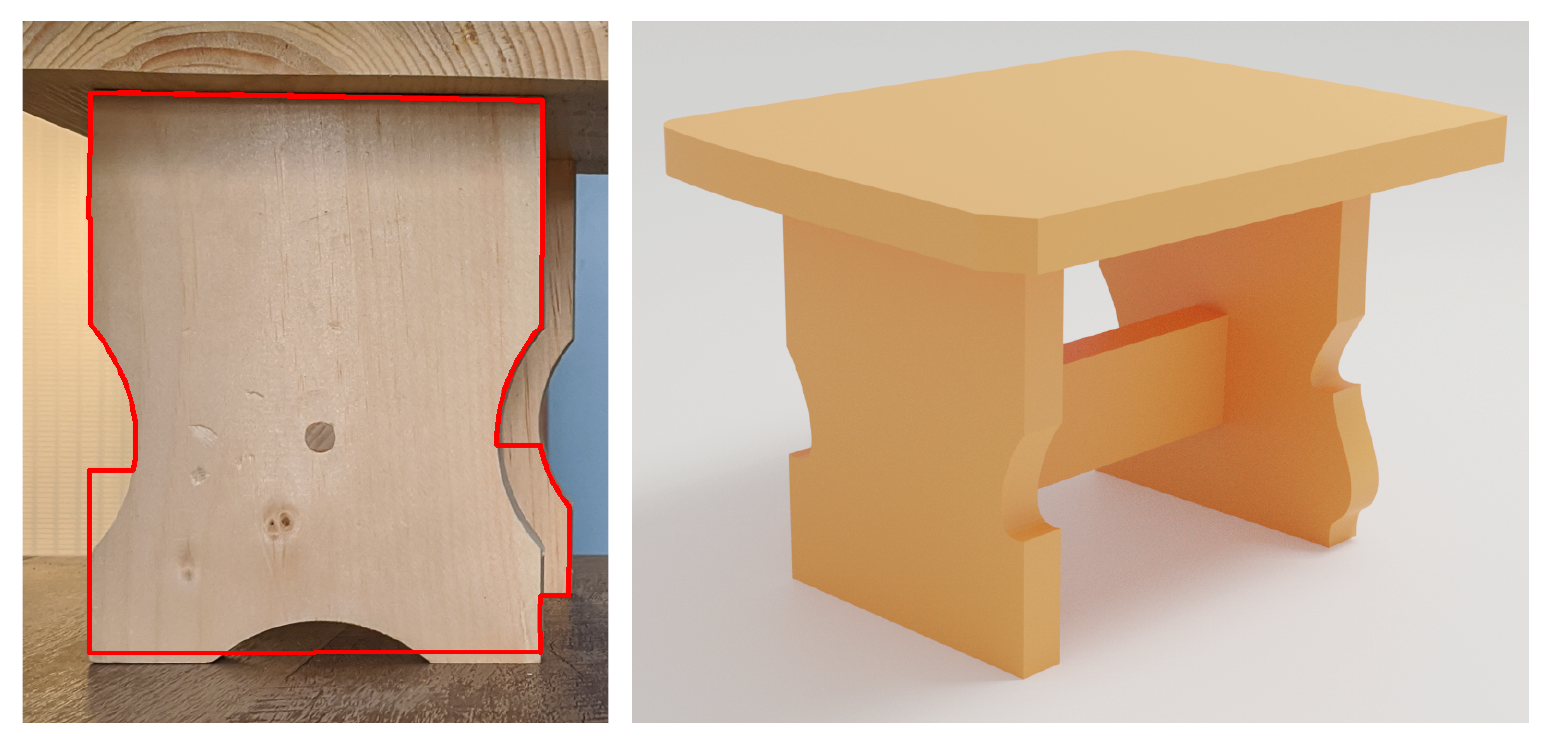}
    \caption{Results from using only one image per part in the segmentation phase. Left: segmentation result for one part, with the segmented shape superimposed in red on the corresponding image. Right: The full result.}
    \label{fig:singleimage}
\end{figure}

\paragraph*{Joint Image-Based Segmentation}
\hle{Figure} \ref{fig:nosegmentation} \hle{shows the result of skipping image segmentation, and directly applying curve fitting to the point set boundaries from Section 4.2.2. Because regions of the model with less visibility, such as cavities and undersides, are often missing from the point cloud, the final result contains gaps. Compared to the full pipeline, we lack the ability to expand part shapes into regions of similar texture, which is a method to close gaps such as this.}

\paragraph*{Using Multiple Views in Segmentation}
\hle{We also show results from using only a single image per part in the segmentation phase in Figure} \ref{fig:singleimage}\hle{. Without multiple views to disambiguate foreground and background, similarly-colored surfaces become erroneously associated with the part shape, leading to an artifact-laden result. In general, material and lighting conditions can make discerning part shapes from certain views difficult; averaging the segmentation energy over multiple reprojected views exploits the view-dependence of pixels not belonging to the part, since similarly-colored background surfaces are less likely to interfere with the same pixels in every view.}


\begin{figure}
    \centering
    \includegraphics[width=0.7\textwidth]{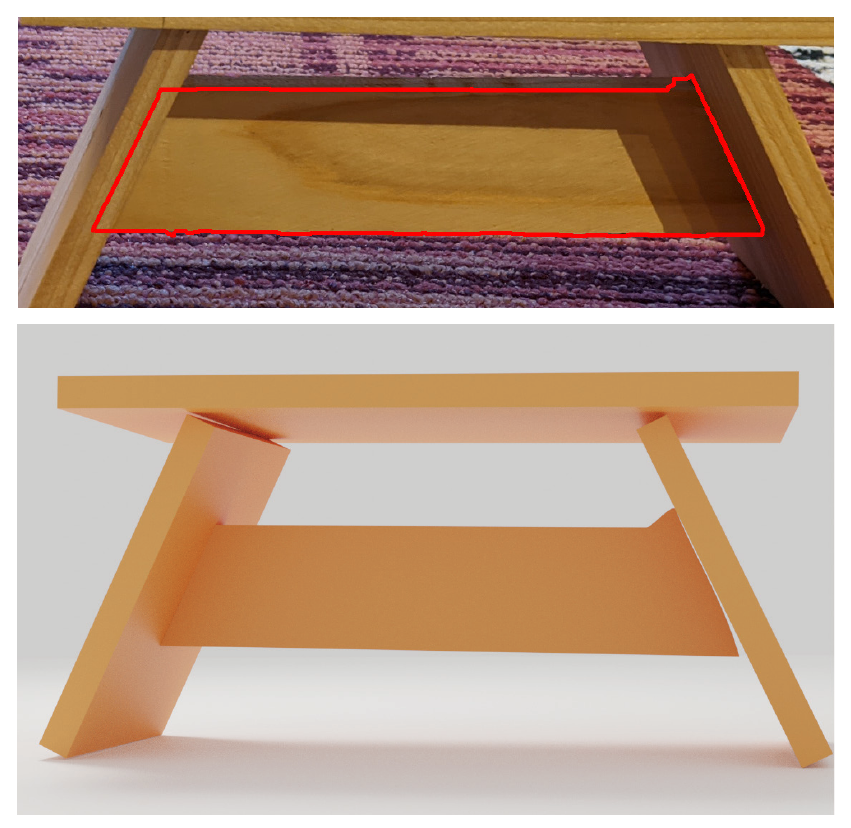}
    \caption{In the presence of some segmentation artifacts (top), unconstrained curve fitting produces an incorrect result (bottom). Note that the supposed contact between the central piece and the right leg is curved.}
    \label{fig:noconstraint}
\end{figure}

\paragraph*{Constrained Curve Fitting}
\hle{The constraints in the curve fitting stage straighten curves in the vicinity of connection contacts, effectively flattening nearby artifacts arising from the segmentation stage (which occur due to the poor visibility at some junctions). In our tilted stool example, one such artifact occurs due to heavy shadowing (see the top of Figure} \ref{fig:noconstraint}\hle{), but our constrained curve fitting gives a clean result (see Figure 8 in the main paper). The bottom of Figure} \ref{fig:noconstraint} \hle{shows the result without considering constraints in the curve fitting stage. The artifact persists; furthermore, the contacts between all parts are imperfect. Amending the contacts purely in a post-process gives rise to new challenges, for without enforcing straight segments in the vicinity of contacts, there may not be a single part of the boundary curve that can be ``snapped'' to the surface. For example, the cut edge on the right of the central part has been chosen as part of a longer, continuous curve.}

\printbibliography
\end{document}